\documentclass[runningheads]{llncs}
\usepackage{graphicx}
\usepackage{amsmath,amssymb} % define this before the line numbering.
\usepackage{color}
\usepackage[width=122mm,left=12mm,paperwidth=146mm,height=193mm,top=12mm,paperheight=217mm]{geometry}
\begin{document}

\newcommand{\etal}{\mbox{\emph{et al.\ }}}
\newcommand{\ie}{\mbox{i.e.\ }}
\newcommand{\eg}{\mbox{e.g.\ }}

% \linenumbers
\pagestyle{headings}
\mainmatter

\title{Chained Predictions Using Convolutional Neural Networks} % Replace with your title

\titlerunning{Chained Predictions Using Convolutional Neural Networks}

\authorrunning{Gkioxari \etal}

\author{Georgia Gkioxari$^1$, Alexander Toshev$^2$, Navdeep Jaitly$^2$ \\
\tt\small{gkioxari@eecs.berkeley.edu, \{toshev, ndjaitly\}@google.com}}
\institute{\centering{$^1$University of California, Berkeley  \hspace{0.5cm} $^2$Google Inc}}

\maketitle

\begin{abstract}
In this work, we present an adaptation of the sequence-to-sequence model for
structured vision tasks. In this model, the output variables
for a given input are predicted sequentially using neural networks. The prediction
for each output variable depends not only on the input but also on the previously 
predicted output variables. The model is applied to spatial
localization tasks and uses convolutional neural networks (CNNs) for processing input images
and a multi-scale deconvolutional architecture for making spatial predictions at each
step. We explore the impact of weight sharing with a recurrent
connection matrix between consecutive predictions, and compare it to a formulation where
these weights are not tied. Untied weights are particularly suited for problems with a
fixed sized structure, where different classes of output are predicted at different steps.
We show that chain models achieve top performing results on human pose estimation from images and videos.
\keywords{Structured tasks, chain model, human pose estimation}
\end{abstract}

%% Main Text
\section{Introduction}

Structured prediction methods have long been used for various vision tasks, such as
segmentation, object detection and human pose estimation, to deal with complicated
constraints and relationships between the different output variables predicted from
an input image.  For example, in human pose estimation the location of one body part
is constrained by the locations of most of the other body parts. Conditional
Random Fields, Latent Structural Support Vector Machines and related methods are
popular examples of structured output prediction models that model dependencies
among output variables.

A major drawback of such models is the need to hand-design the structure of the model in order to capture important problem-specific dependencies amongst
the different output variables and at the same time allow for tractable inference.
For the sake of efficiency, a specific form of conditional independence amongst
output variables is often assumed. For example, in human pose
estimation, a predefined kinematic body model is often used to assume that each
body part is independent of all the others except for the ones it is attached
to. 

To alleviate some of the above modeling simplifications, structured prediction problems
have been solved with sequential decision making, where all earlier predictions
influence later predictions. The SEARN algorithm~\cite{daume2009search} introduced a very general formulation
for this approach, and demonstrated its application to various natural language
processing tasks using losses from binary classifiers. A related model 
recently introduced, the sequence-to-sequence model, has been applied to various
sequence mapping tasks, such as machine translation, speech recognition and image
caption generation~\cite{seq2seq,chan2015listen,vinyals2015show}. In all these models
the output is a sentence - where the words of the sentence are predicted in a first
to last order. This model maximizes the log probability for output sequence 
conditioned on the input, by decomposing the probability of an output sequence 
with the multiplicative chain rule of probability; at each index of the output, 
the next prediction is made
conditioned on all previous outputs and the input. A recurrent neural network
is used at every step of the output and this allows parameter sharing across
all the output steps.

\begin{figure}[t!]
\centering
\includegraphics[width=12cm]{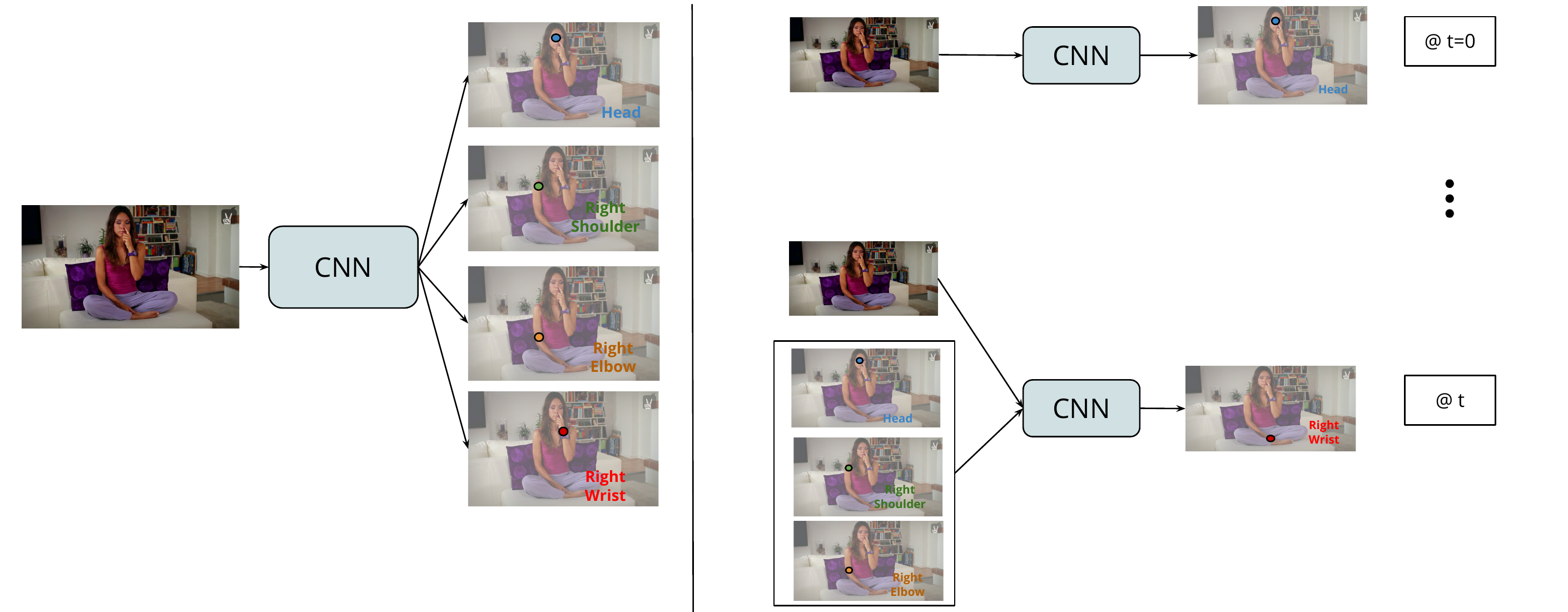}
\caption{A description of our model for the task of body pose estimation compared to pure feed forward nets. \textbf{Left}: Feed forward networks make independent predictions for all body parts simultaneously and fail to capture contextual cues for accurate predictions. \textbf{Right}: Body parts are predicted sequentially, given an image and all previously predicted parts. Here, we show the chain model for the prediction of \emph{Right Wrist}, where predictions of all other joints in the sequence are used along with the image.}
\label{fig:fig1}
\end{figure}

In this paper we borrow ideas from the above sequence-to-sequence model and propose
to extend it to more general structured outputs encountered in computer vision -- human 
pose estimation from a single image and video. The contributions of this work are as follows:
\begin{itemize}
\item A \textit{chain model for structured outputs}, such as human pose estimation. The
body part locations are predicted sequentially, where the prediction of each body
part is dependent on \textit{all previously predicted} body parts (See Fig.~\ref{fig:fig1}).
The model is formulated using a neural network in which the feature extraction and prediction
models are learned end-to-end. Since we apply the model to spatial labelling tasks
we use convolutional neural networks in both the inputs and outputs. The output
convolutional neural networks is a multi-scale deconvolution that we call \textit{deception}
because of its relationship to deconvolution~\cite{dosovitskiy2015learning,long2015fully} and inception models~\cite{inception}.

\item We demonstrate \textit{two formulations of the chain model} - one without weight sharing
between different predictors (poses in images) to allow semantic-specific flow of information and the other with weight sharing to enforce recurrence in time
(poses in videos). The latter model is a RNN similar to the sequence-to-sequence model.
\end{itemize}

The above model achieves top performing results on the MPII human pose dataset -- 86.1\% PCKh. We achieve state-of-the art performance for pose estimation on the PennAction video dataset -- 91.8\% PCK.

\section{Related Work}

\paragraph{Structured output prediction as sequence prediction.}
The use of sequential models for structured predictions is not new. The
SEARN algorithm~\cite{daume2009search} laid down a broad framework for such models in which a sequence
of actions is generated by conditioning the next action on previous actions
and the data. The optimization method proposed
in SEARN is based on iterative improvement over policies using reinforcement learning.

A similar class of models are the more recent sequence-to-sequence models~\cite{seq2seq,attention}
that map an input sequence to an output sequence of fixed vocabulary. The models produce
output variables, one at a time, conditioned on inputs and previous output variables.
A next-step loss function is computed at each step, using a recurrent neural network.
Sequence-to-sequence models have been shown to be very effective at a variety of 
language tasks including machine translation~\cite{seq2seq},
speech recognition~\cite{chan2015listen}, image captioning~\cite{vinyals2015show} and
parsing~\cite{vinyals2015grammar}. In this paper we use the same idea of chaining
predictions for structured prediction on two vision problems - human pose estimation in
individual frames and in video sequences.  However, as exemplified
in the pose estimation case, since we have a fixed output structure we are not
limited to using recurrent models.

In the pose prediction problem, we used a fixed ordering of joints, that
is motivated by the kinematics of the human body. Prior work in
sequential modelling has explored the idea of choosing the best
ordering for a task~\cite{ordermatters,goldberg2010efficient,dagger}. For example,
Vinyals et al.~\cite{ordermatters} explored this question and found that for some problems,
such as geometric problems, choosing an intuitive
ordering of the outputs results in slightly better performance. However for
simpler problems most orderings were able to perform equally well. For our problem,
the number of joints being predicted is small, and tree based ordering of joints from
head to torso to the extremities seems to be the intuitively correct ordering. 
\paragraph{Human pose estimation}
Human pose estimation has been one of the major playgrounds for
structured prediction models in computer vision. Historically, most of the research
has focused on graphical models, starting with tree-based 
decompositions~\cite{felzenszwalb2005pictorial,ramanan2006learning,andriluka2009pictorial,eichner2009better}
motivated by kinematic models of the human body. 

Many of these models assume conditional independence of a body part from all other 
parts except the parent part as defined by the kinematic body model (see pictorial structure model~\cite{felzenszwalb2005pictorial}). 
This simplification comes at a performance cost and has been addressed in various ways: 
mixture model of 
parts~\cite{yang2011articulated}; mixtures of full body models~\cite{Johnson11}; 
higher-order spatial relationships~\cite{tian2012exploring}; image dependent pictorial
structures~\cite{wang2013beyond,modec13,pishchulin2013poselet,Karlinsky2010}. Like these above approaches,
we assume an order among the body parts. However, this ordering is used only to decompose
the joint probability of the output joints into a particular ordering of variables in
the chain rule of probability, and not to make assumptions about the structure of the
probability distribution. Because no simplifying assumptions are made about the 
joint distribution of the output variables it leads to a more expressive model, as
exemplified in the experimental section. The model is only constrained by the
ability of neural networks to model the conditional probability distributions
that arise from the particular ordering of the variables chosen. In addition, the correlations among parts are learned through a set of non-linear operations instead of imposing binary term constraints on hand-designed image features (\eg RGB values, location) as done in CRFs.

It is worth noting that there have been models for pose estimation where parts are
sequentially refined~\cite{toshev2014,ramakrishna2014pose,IEF2015human,wei16}. In these
models an initial prediction is made of all the parts; in subsequent steps,
all part predictions are refined based on the image and earlier part predictions.
However, note that the predictions are initially independent of each other.

\section{Chain Models for Structured Tasks}
\label{sec:method}

\begin{figure}[t!]
\includegraphics[width=6cm]{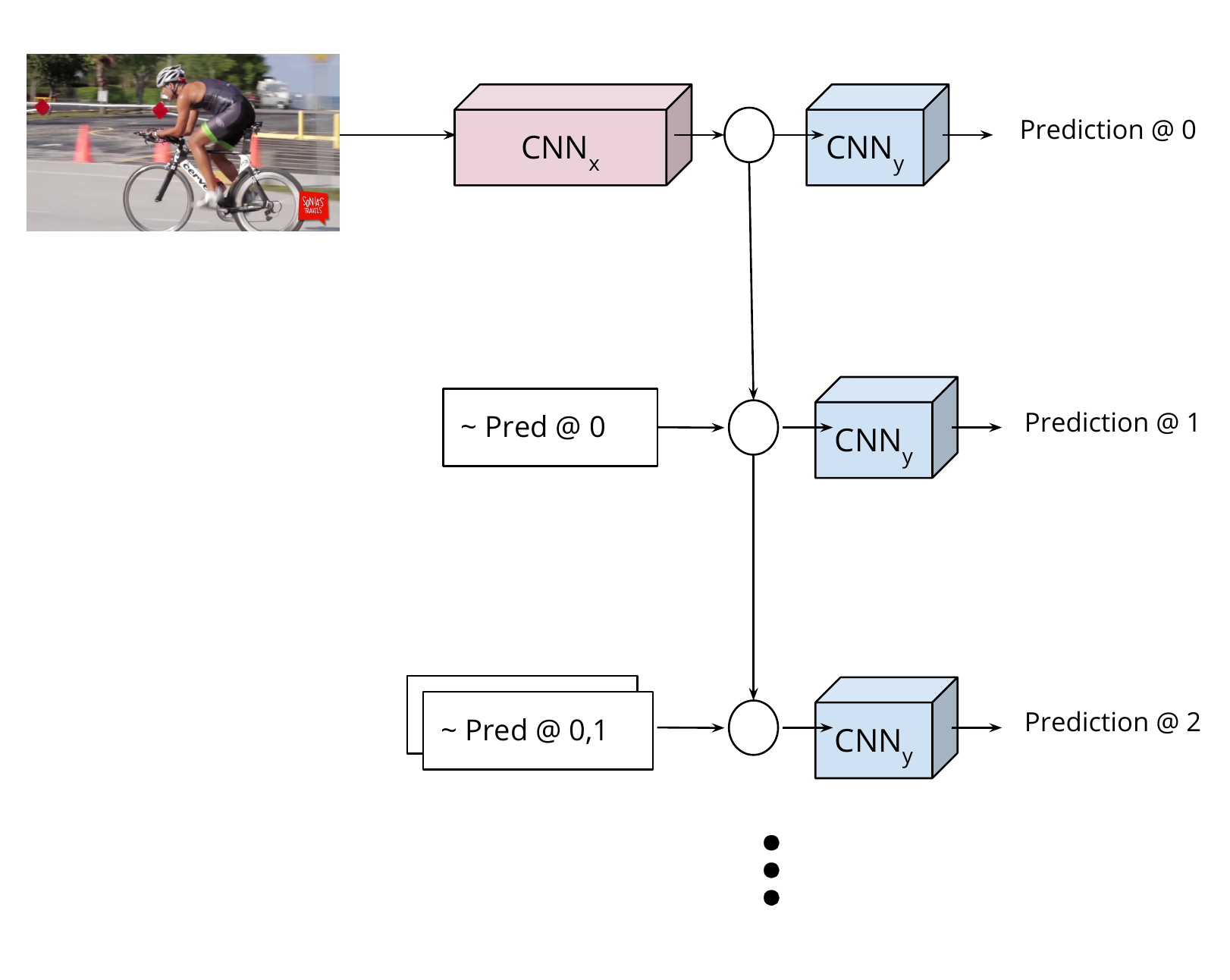}
\includegraphics[width=6.2cm]{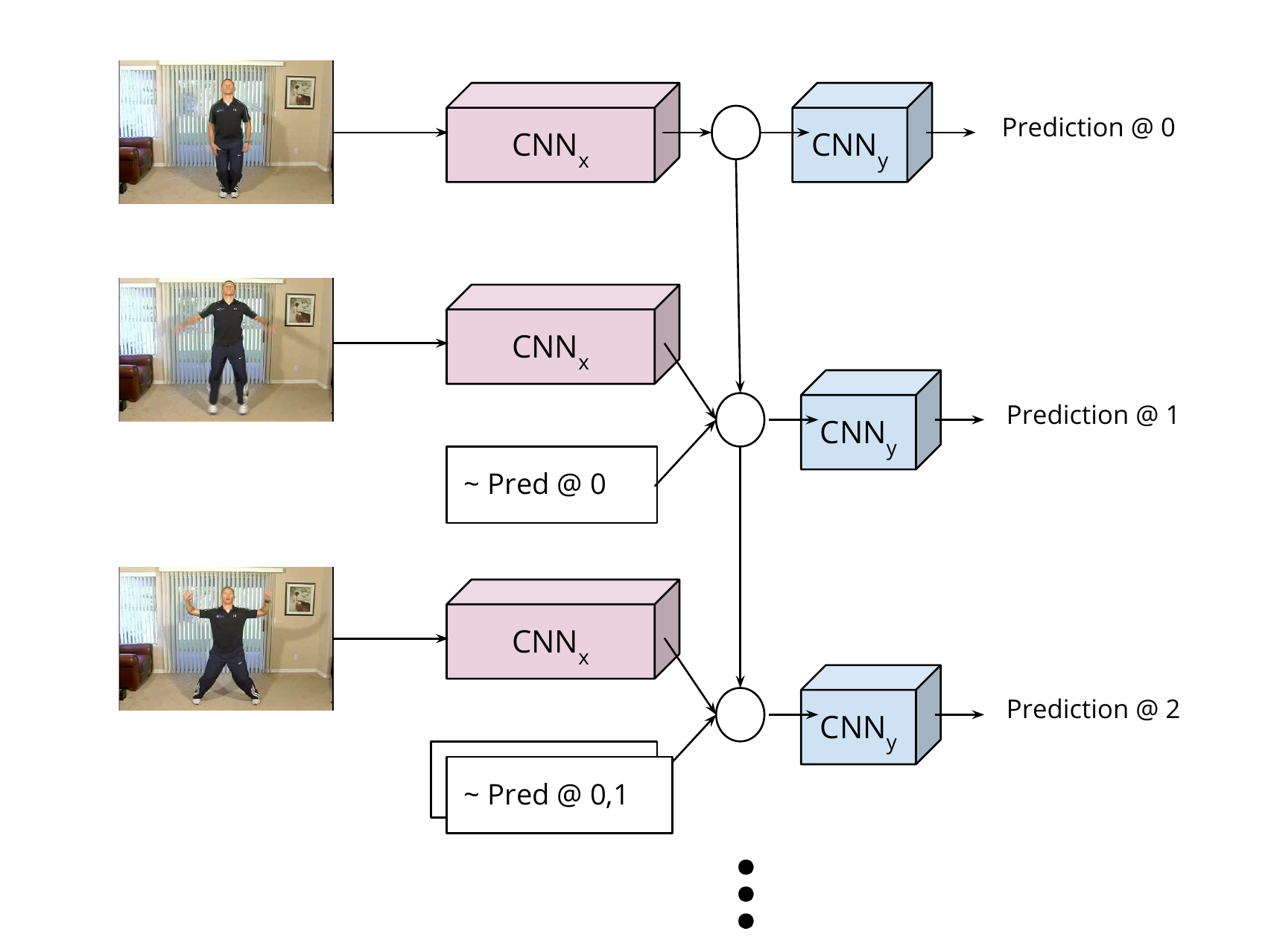}
\caption{A visualization of our chain model. \textbf{Left:} single image case. \textbf{Right:} video case. In both cases, an image is encoded with a
CNN ($\textrm{CNN}_x$). At each step, the previous output variables are combined with the hidden state, through the sequential modules. A CNN decoder ($\textrm{CNN}_y$) makes predictions each step, $t$. There are two differences between the two cases: (i) for video $\textrm{CNN}_x$ receives at each step a frame as an input, while for single image there is no such input; (ii) for video $\textrm{CNN}_y$ share parameters across steps, while for single image the parameters are untied. }
\label{fig:models}
\end{figure}

Chain models exploit the structure of the tasks they are designed to tackle by sequentially predicting their outputs.
To capture this structure each output prediction is conditioned on all outputs predicted already. This philosophy has been
exploited in language processing where sentences, expressed as word sequences, need to be predicted~\cite{seq2seq,attention}
from inputs.  In recent automatic image captioning work~\cite{vinyals2015show,attentioncaption}, for example, a
sentence $Y$ is generated from an image $X$ by maximizing the likelihood $P(Y \, | \, X)$. The chain rule is applied,
consecutively to model each output $Y_t$ (here a word) given the image $X$ and all the previous outputs $Y_{<t}$ in
the output sequence.

In computer vision, recognition problems, such as segmentation, detection and pose estimation, demonstrate rich structure
with complex dependencies. In this work, we model this structure with a simple and efficient recognition machine
that makes little to no assumptions about the structure, other than the ability of a neural network to model complex,
incremental conditional distributions.

Mathematically, let $Y=\{Y_t\}^{T-1}_{t=0}$ be the $T$ objects to be detected. For example,
for the pose prediction problem, $Y_t$ is the location of the $t$-th body part. In video prediction problems, $Y_t$
is the location of an object in the $t$-th frame of a video.  Using the chain rule we decompose
$P(Y=y \, | \, X)$ as follows:

\begin{align}
P(Y=y \, | \, X) = P(Y_0=y_0 \, | \, X) \prod_{t=1}^{T-1} P(Y_t=y_t \, | \, X,y_0,...,y_{t-1})
\label{eq:chain}
\end{align}

From the above equation, we see that the likelihood of assigning value $y_t$ to the $t$-th variable is given by $P(Y_t=y_t \, | X,y_0,...,y_{t-1})$, and
depends on both the input $X$ as well as the assignment of previous variables. In this work, we model the likelihood $P(Y_t=y_t \, | X,y_0,...,y_{t-1})$
with a convolutional neural network (CNN).
The direct dependence of the current prediction on the ground truth values of previous variables allows for the model to capture all
necessary relationships without making any assumption about the joint distributions of all the variables, other
than assuming that each successive conditional distribution, $P(Y_t=y_t \, | X,y_0,...,y_{t-1})$, can be computed
with a neural network.

\subsection{Chain Models for Single Images}
In the case of single images, the input $X$ is the image while the $t$-th variable $Y_t$ can be, for example, the location
of the $t$-th object in image $X$ (see Fig.~\ref{fig:models}).

The probability of each step in the decomposition of Eq.~(\ref{eq:chain}) is defined through a hidden state $h_t$ at step $t$, which carries information about the input as well as states at previous steps. In addition it incorporates the values $y_{<t}$ from previous steps. The final probability
for variable $Y_t$ is computed from the hidden state:

\begin{align}
&h_t = \sigma( w_{t}^{h} \ast h_{t-1} + \sum_{i=0}^{t-1} w_{i,t}^{y} \ast e(y_i)) \label{eq:hidden} \\
&P(Y_t=y_t \, | \, X,y_0,...,y_{t-1}) = \textrm{Softmax}(m_t(h_t)) \label{eq:prob} 
\end{align}

In the above equation, the previous variables are first transformed through a full neural net $e(\cdot)$.
Parameters $w_{t}^{h}$ and $w_{i,t}^{y}$ then linearly transform the previous hidden state and a function of
previous output variables, $e(\cdot)$, and a non-linearity $\sigma$ is then applied to each dimension of
this output. The nonlinearity $\sigma$ of choice is a Rectified Linear Unit. Finally, $\ast$ denotes
multiplication. In image applications, however, the hidden state $h$ can be a feature map and the
prediction $y$ a location in the image. In such cases, $\ast$ denotes convolution and $e$ is a CNN.
Note that, as long as we feed in just the last variable $y_{t-1}$ in this
equation, the recurrent equation insures that we condition on the entire history of joints. However feeding in
more of the previous joints makes it easier for the model to learn the conditional distributions directly. 
In the computation of the conditional probability of $y_t$ from $h_t$ we use another neural net $m_t$, which
produces scores for potential object location. By applying a softmax function over these scores we convert
them to a probability distribution over locations.

The initial state $h_0$ is computed based solely on the input $X$: $h_0 = \textrm{CNN}(X)$.

This formulation is reminiscent of recurrent networks (RNNs), the equations define how to transform a
state from one step to the next. We differ, however, from RNNs in one important aspect, the parameters in
Eq.~(\ref{eq:hidden}-\ref{eq:prob}) are not necessarily tied. Indeed, parameters
$w_{t}^{h}$ and $w_{i,t}^{y}$ are indexed by the step. This design choice is appropriate for tasks such
as human pose estimation where the number of outputs $T$ is fixed and where each step is different
from the rest. In other applications, \eg video, we tie these parameters:
$w_{t}^{h} = w_{0}^{h}$ and $w_{i,t}^{y}=w_{i,0}^{y}$, $\forall i,t$.

\subsection{Chain Models for Videos}
For videos, the input is a sequence of images $X=\{X_t\}_{t=0}^{T-1}$ (Fig.~\ref{fig:models}). Predictions are made at each step, as the images are fed in. At each step $t$, we make predictions for the image $X_t$ at that step, using the past images, and the past output variables.
Thus, we modify the equation for the hidden state as follows:
\begin{align}
&h_t  = \sigma( w_{t}^{h} \ast h_{t-1} + \textrm{CNN}(X_t) + \sum_{i=t-T_H}^{t-1} w_{t-i,t}^{y} \ast e(y_i))
\label{eq:recur_vid}
\end{align}
where we add features extracted from image $X_t$ using a CNN. The final probability is computed as in Eq.~(\ref{eq:prob}).

In videos we often need to predict the same type of information at each step, e.g.~location of all body
joints of the person in the current frame. As such, the predictors can have the same weights. Thus,
we tie the parameters $w_{t}^{h}$, $ w_{i,t}^{y}$, and $m_t$ together, which results in a convolutional RNN.

As before, the connections from hidden state at the previous step guarantees that the prediction at
each time step uses output variables from all previous steps, as long as the previous output variable $Y_{t-1}$
is fed in at time $t$.  However, feeding in a larger time horizon $T_{H}$ leads to an easier
learning problem.

\subsection{Improved Learning with Scheduled Sampling}
\label{sect:training}
So far, we have described the method as using the input and only ground truth values of the
previous output variables when making a prediction for the next output variable. However,
it has previously been observed that for sequence-to-sequence models overfitting can be mitigated
by probabilistically substituting ground truth values of previous output variables with samples
from the probability distribution predicted by the model~\cite{sched_sample}. One challenge
that arises in this is that, at the start of the training, the predicted probability distributions are wildly
inaccurate and thus, feeding in samples from the distribution is counter-productive.
The authors of~\cite{sched_sample} propose a method, called {\it scheduled sampling}, that uses an annealing schedule that feeds in only the ground truth outputs at the start of the training and increases
the rate of sampling from the predictions of the model towards the end of the training. We
use the idea of scheduled sampling in our paper and find that it leads to improved results.

\section{Experimental Evaluation}
\label{sec:pose}
To evaluate the proposed model, we apply it on human pose estimation, which is challenging and of great interest due to the complex relationship among body parts. In the single image case, we use the chain model to capture the structure of pose in space, \ie how the location of a part influences others. For the videos, our model captures the constraints and dynamics of the body pose in time.
\subsubsection{Tasks and Datasets}
For our single image experiments we use the MPII Human Pose dataset \cite{andriluka14cvpr}, which consists of about 40K instances of people performing various actions. All frames come with a maximum of 16 annotated joints (\eg \emph{Top Head}, \emph{Right Ankle}, \emph{Left Knee}, etc.). For the task of pose estimation in video we use the Penn Action dataset \cite{penn-action}, which consists of 2326 video sequences of people performing various sports. All frames come with a maximum of 13 annotated joints. During evaluation, if a joint prediction lies within a predefined distance, proportional to the size of the person, from the ground truth location it is counted as a correct detection. This metric is called PCK~\cite{yang2012articulated,andriluka14cvpr}.

\begin{figure}[b!]
\centering
\includegraphics[height=6cm]{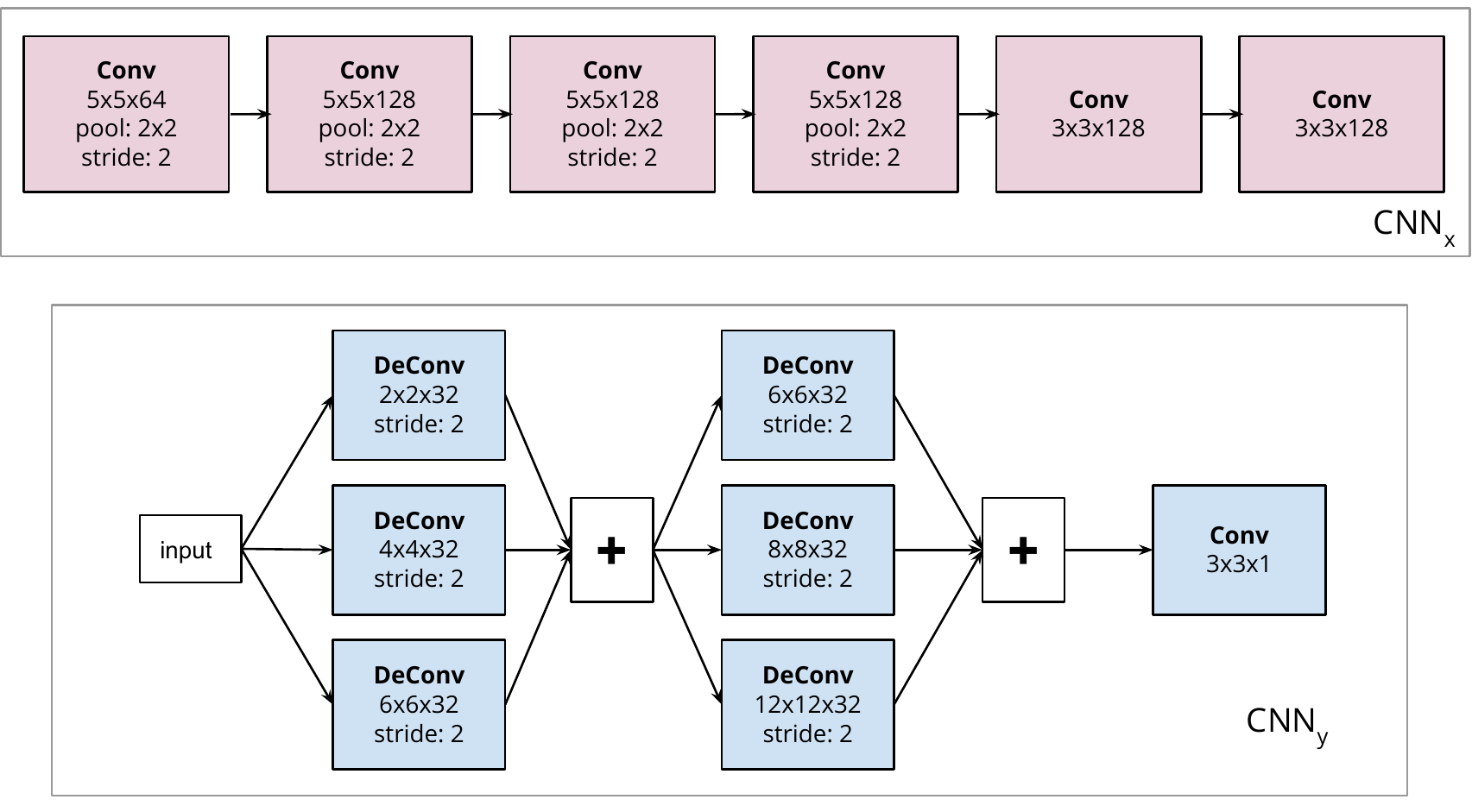}
\caption{Description of the components of our network, $\textrm{CNN}_x$ and $\textrm{CNN}_y$. Each box represents a convolutional or deconvolutional layer, where $w \times h \times f$ denotes the width $w$, the height $h$ of the filters and $f$ denotes the number of filters. In each layer the filter is applied with stride 1 if not noted otherwise. Finally, in each layer after the filtering operation a ReLU and batch normalization are applied.}
\label{fig:network}
\end{figure}

Our model is illustrated in Fig.~\ref{fig:models}. We experiment with two choices for $\textrm{CNN}_x$, the network which encodes the input image. First, a shallow CNN which consists of six layers each followed by a rectified linear unit~\cite{nair2010rectified} and Batch Normalization~\cite{ioffe2015batch}. The first four layers include max pooling with stride 2, leading to an effective stride of 16. This network is described in Fig.~\ref{fig:network}. Second, we experiment with a deeper network of identical architecture to inception-v3~\cite{SzegedyVISW15}. We discard the last convolutional layer of inception-v3 and connect the output to $\textrm{CNN}_y$. 

The $\textrm{CNN}_y$ network decodes the hidden state to a heatmap over possible locations of a single body part. This heatmap is converted to a probability distribution over locations using a softmax. The network consists of two towers of deconvolutional layers each of which increases the width and height of the feature maps by a factor of 2. Note that the deconvolutional towers are multi-scale - in one layer, different filter sizes are used and combined together. This is similar to the inception model~\cite{inception}, with the difference that here it is applied with the deconvolution operation, and hence we call it \textit{deception}.

\subsection{Pose Estimation From a Single Image}
In this application case, we use the chain model to predict the joints sequentially. The sequence with which the joints are processed is fixed and is motivated by the marginal distributions of the joints. In particular, we sort the joints in descending order according to the detection rates of an unchained feed forward net. This allows for the easy cases to be processed first (e.g. \emph{Torso}, \emph{Head}) while the harder cases (e.g. \emph{Wrist},  \emph{Ankle}) are processed last, and as a result use the contextual information from the joints predicted before them. 

\subsubsection{Inference}
At test time, we use beam search to infer the optimal location of the joints. Note that exact inference is infeasible, due to the
size of the search space (a total of $(HW)^T$ possible solutions, where $H \times W$ is the size of the prediction heatmap and $T$ are the number of joints). At each step $t$, the best $B$ predictions are stored, where each prediction is the sequence of the first $t$ joints.
The quality of a full body pose prediction is measured by its log-probability, which is the sum of the log-probabilities corresponding to the individual joint predictions.

An exact implementation of chain rule conditions on predictions made at every step. Alternatively, one could skip the non-differentiable sampling operation and use the probability distributions directly. Even though this is not an exact application of the chain rule, it allows for the gradients to flow back to the output of each task. We found that this approximation led to very similar performance - it slowed down training time by a factor of 3 and sped up inference by a factor of $B$.

\setlength{\tabcolsep}{4pt}
\begin{table}[b!]
\begin{center}
	\caption{PCKh performance on the MPII validation set. Rows 1 and 2
		show results for 9-layered CNN models, with multi-scale (deception) and single scale deconvolutions. Row 3
		show results for a 24-layer model with deception, but without chained outputs. Row 4 shows results for our
		chain model with comparable depth and number of parameters as the 24-layer model, but with chained predictions.
		We observe clear improvement over the baselines. The performance is further improved using multiple crops of the 
		input at test time, at row 5. Row 6 shows the performance of the oracle, where the
		correct values of previous output is fed into the network at each step. Row 7 and 8 show the performance for a base and chain model when inception-v3, pre trained on ImageNet, is used as the encoder network. Using a deeper architecture leads to substantially improved results across all joints.}
\label{table:base}
\begin{tabular}{l|c|c|c|c|c|c|c|c||c}
PCKh  (\%)                                                              & Torso         & Head          & Shldr      & Elbow         & Wrist         & Hip           & Knee          & Ankle         & Mean       \\
\hline
\hline
\begin{tabular}[c]{@{}l@{}}Base Network  \end{tabular}                            &    86.8    &    91.9     &    85.8      &     74.5    &    69.0      &     71.1     &    61.4     &      50.6      &  73.9 \\
\hline
\begin{tabular}[c]{@{}l@{}}Base Net. w/ \\ single deconv. \end{tabular}    &     86.0    &    91.7     &    85.1      &      72.9     &     68.0      &    69.4    &    59.7    &     48.5     &  72.6 \\
\hline
\begin{tabular}[c]{@{}l@{}}Very Deep \\ Base Network\end{tabular}         &    88.1    &     92.0    &      86.1    &     74.1      &    67.7       &   73.7     &   64.7      &      58.0    &  75.6 \\
\hline
\begin{tabular}[c]{@{}l@{}}Chain Model\end{tabular}                     & 86.8       &  93.2  &  88.3  &  79.4  &  74.6  & 77.8  &  71.4  & 65.2 & 79.6 \\
\hline
\begin{tabular}[c]{@{}l@{}}Chain Model \\ w/ multi-crop \end{tabular}   & 88.7       &  94.4  &  90.0  &  82.6  &  78.6  & 80.2  &  74.8  & 68.4 & 82.2 \\
\hline
\begin{tabular}[c]{@{}l@{}}Oracle \\ Chain Model\end{tabular}              &     87.2     &     95.9     &      93.4    &     83.3    &     82.3     &    95.2      &    77.6      &     72.3    &  85.9  \\
\hline
\hline
\begin{tabular}[c]{@{}l@{}}Inception \\ Base Network\end{tabular}   & 91.1 & 95.0 & 90.2 & 81.0 & 77.4 & 77.2 & 73.7 & 64.6 & 81.3 \\
\hline
\begin{tabular}[c]{@{}l@{}}Inception \\ Chain Model\end{tabular} & 91.7 & 95.7 & 92.2 & 85.3 & 82.2 & 82.9 & 80.0 & 72.4 & 85.3  
\end{tabular}\end{center}
\end{table}
\setlength{\tabcolsep}{1.4pt}
 
\subsubsection{Learning details}
We use an SGD solver with momentum to learn the model parameters by optimizing the loss. The loss for one image $X$ is defined as the sum of losses for individual joints. The loss for the $k$-th joint is the cross entropy between the predicted probability $P_k$ over locations of the joint and the ground-truth probability $P_{k}^{\textrm{gt}}$. The former is defined based on the heatmap $h_k$ output by $\textrm{CNN}_y$ for the $k$-th joint: $P_k(x,y) = \frac{e^{h_k(x,y)}}{\sum_{(x',y')} e^{h_k(x',y')}}$. The latter is defined based on a distance $r$ -- all locations within radius $r$ of the ground-truth joint location are assigned same nonzero probability $P_{k}^{\textrm{gt}}(x,y)=1/N$, all other locations are assigned probability $0$. $N$ is a normalizer guaranteeing $P_{k}^{\textrm{gt}}$ is a probability.

The final loss for $X$ reads as follows:
\begin{align}\label{eq:single_image_loss}
L(\{h_k\}_{k=0}^{T-1}) = \sum_{k=0}^{T-1} \sum_{(x,y)}P_{k}^{\textrm{gt}}(x,y)  \log P_k(x,y)
\end{align}

We use batch size of 16; initial learning rate of 0.003 that was decayed every 100K steps (50K for the inception model); radius of $r=0.01 \times (W + H) / 2$. The model was trained for 120K iterations (55K for the inception model). Our images are rescaled to $224\times224$ ($299\times299$ for the inception model). The weights of the network are initialized by sampling from a normal distribution of zero mean and 0.01 standard deviation. For the inception model, we initialize the weights of $\textrm{CNN}_x$ with weights from an ImageNet model.

\subsubsection{Results}
\label{sec:single_results}
Table~\ref{table:base} shows the PCKh performance on the MPII validation set of our chain model and our baseline variants.

{\it Rows 1, 2 \& 3} show the performance of pure feed forward networks for the task in question. The 1st row shows the performance
of a 9-layer network, shallow $\textrm{CNN}_x$ + $\textrm{CNN}_y$, which we call base network. The 2nd row is a similar network, where each
deconvolutional tower, which we call {\it deception}, in $\textrm{CNN}_y$ is replaced by a single deconvolution. The difference in
performance shows that multi-scale deconvolutions lead to a better and very competitive baseline.
Finally, the 3rd row shows the performance of a very deep network
consisting of 24 layers. This network has the same number of parameters and the same depth as our chain model and serves
as the baseline which we improve upon using the chain model.

{\it Row 4} shows the performance of our chain model. This model improves significantly over all the baselines.
The biggest gains are observed for \emph{Wrists} and \emph{Ankles}, which is a clear indication that conditioning on
the predictions of previous joints provides cues for better localization. 

{\it Row 5} shows the performance of the chain model with multi-crop evaluation, where at test time we average the predictions from flipping and jittering of the input image.

{\it Row 6} shows the performance of an \emph{oracle} chain model. For this model, at each step $t$ we use the oracle (ground truth) locations of all previous joints. This model is an estimate of the upper bound performance of our chain model, as it predicts the location of a joint given perfect knowledge of the location of all other joints which precede it in the sequence.

{\it Row 7} shows the performance of the inception base network, $\textrm{CNN}_x$ + $\textrm{CNN}_y$, where $\textrm{CNN}_x$ is the inception-v3 \cite{SzegedyVISW15}. We observe significant gains when using the inception-v3 architecture compared to a shallower 6-layer network for the encoder network, at the expense of more computations.

{\it Row 8} shows the performance of the inception chain model. For both the inception base and chain model we use multi-crop evaluation. In both cases, the inception-v3 parameters were initialized with weights from an ImageNet model. The inception chain model leads to significant gains compared to its base network (row 7). The improvements are more evident for the joints of \emph{Wrist, Knee, Ankle}.

\subsubsection{Error Analysis}
Digging deeper into the models, we perform an error analysis for the base network $\textrm{CNN}_x$ + $\textrm{CNN}_y$, the very deep network and our chain model. For this analysis, the 6-layer encoder network $\textrm{CNN}_x$ is used for all models. Similar to \cite{hoiem2012diagnosing}, we categorize the erroneous predictions into the three distinct classes: a) localization error, \ie the prediction is within $[\alpha, \beta]\times HeadSize$ of the true location, b) confusion with other joints, \ie the prediction is within $\alpha \times HeadSize$ of a different joint, and c) confusion with the background, \ie the prediction lies somewhere else in the image. According to PCKh, a prediction is correct if it falls within $0.3\times HeadSize$. We set $\beta=0.5$

\begin{figure}[t!]
\centering
\includegraphics[width=2.8cm]{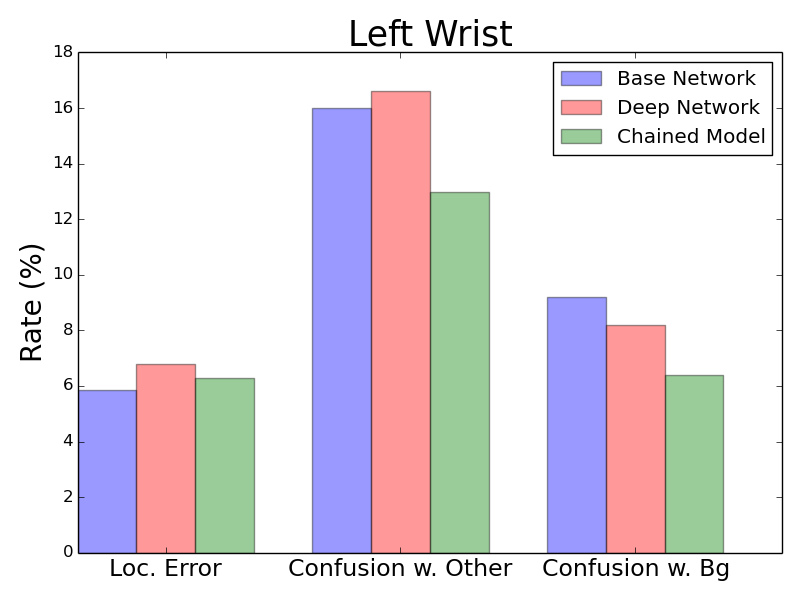}
\includegraphics[width=2.8cm]{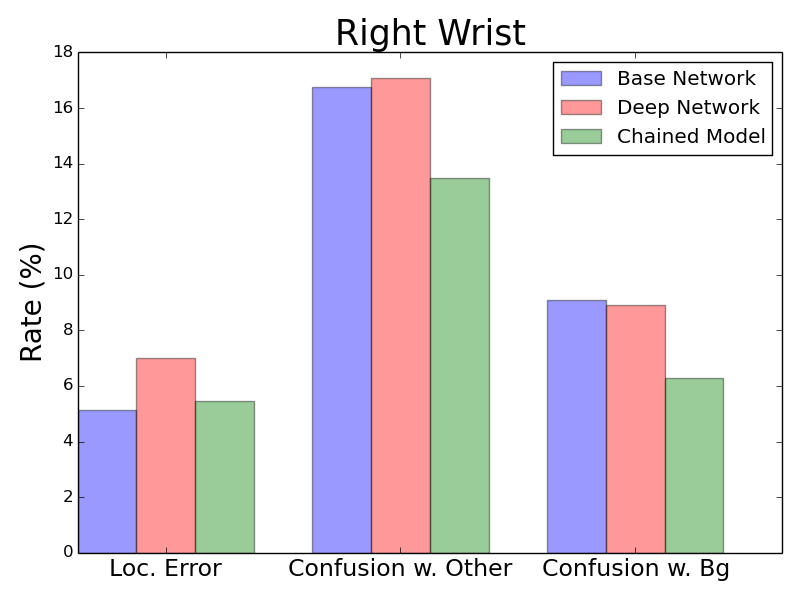}
\includegraphics[width=2.8cm]{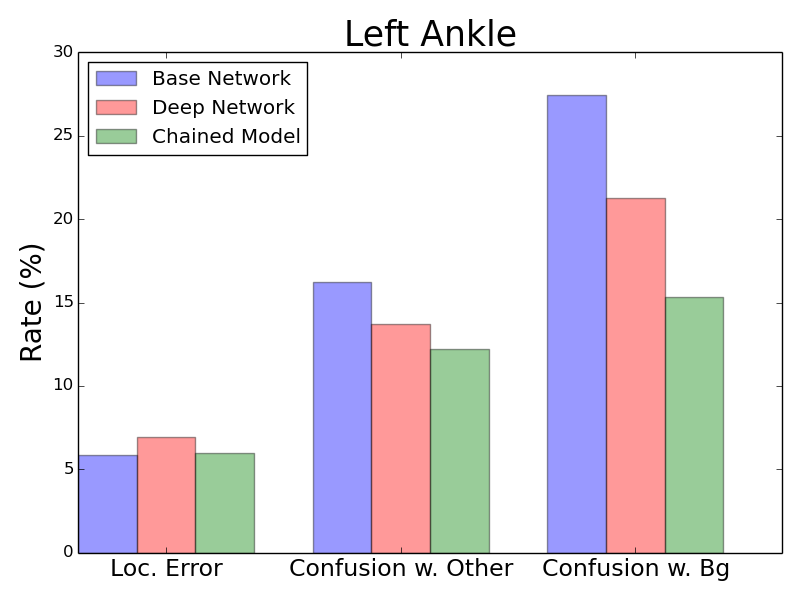}
\includegraphics[width=2.8cm]{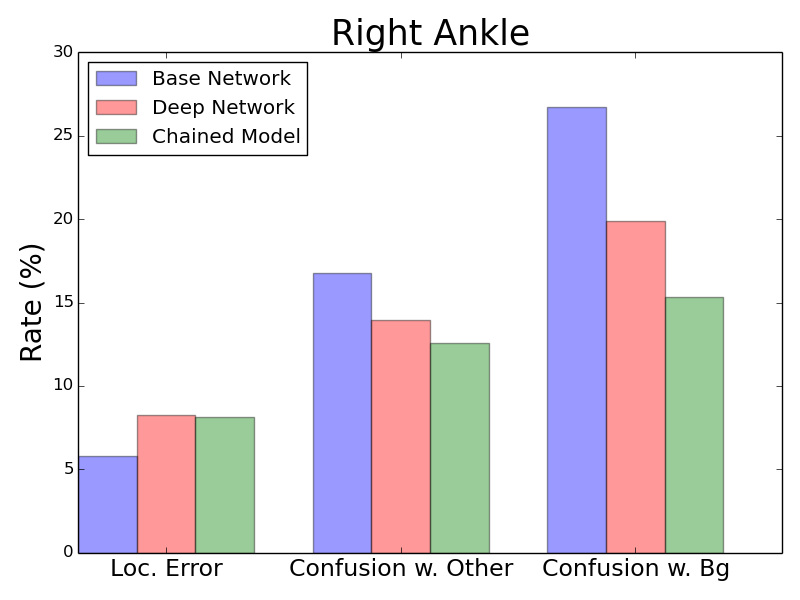}
\caption{Error analysis of the predictions made by the base network (blue), the very deep model (red) and our chain model (green), for \emph{Wrist} and \emph{Ankle}. Each figure shows the error rates, categorized in three classes, localization error, confusion with other joints and confusion with the background.}
\label{fig:error_bars}
\end{figure}

Fig.~\ref{fig:error_bars} shows the error analysis for the hardest joints, namely \emph{Wrist} and \emph{Ankle}. Each plot consists of three sets of bars, the rates for error localization, confusion with other joints and confusion with background. According to the plots, the chain model reduces the misses due to confusion with other joints and the background. For \emph{Wrists}, the confusion with other joints is the dominating error mode, and further analysis shows that the main source of confusion comes mainly from the opposite wrist and then the nearby joints. For \emph{Ankles}, the biggest error mode comes from confusion with the background, which is not surprising since lower legs are usually heavily occluded and lack strong appearance cues. 

\begin{figure}[b!]
\centering
\includegraphics[height=4cm]{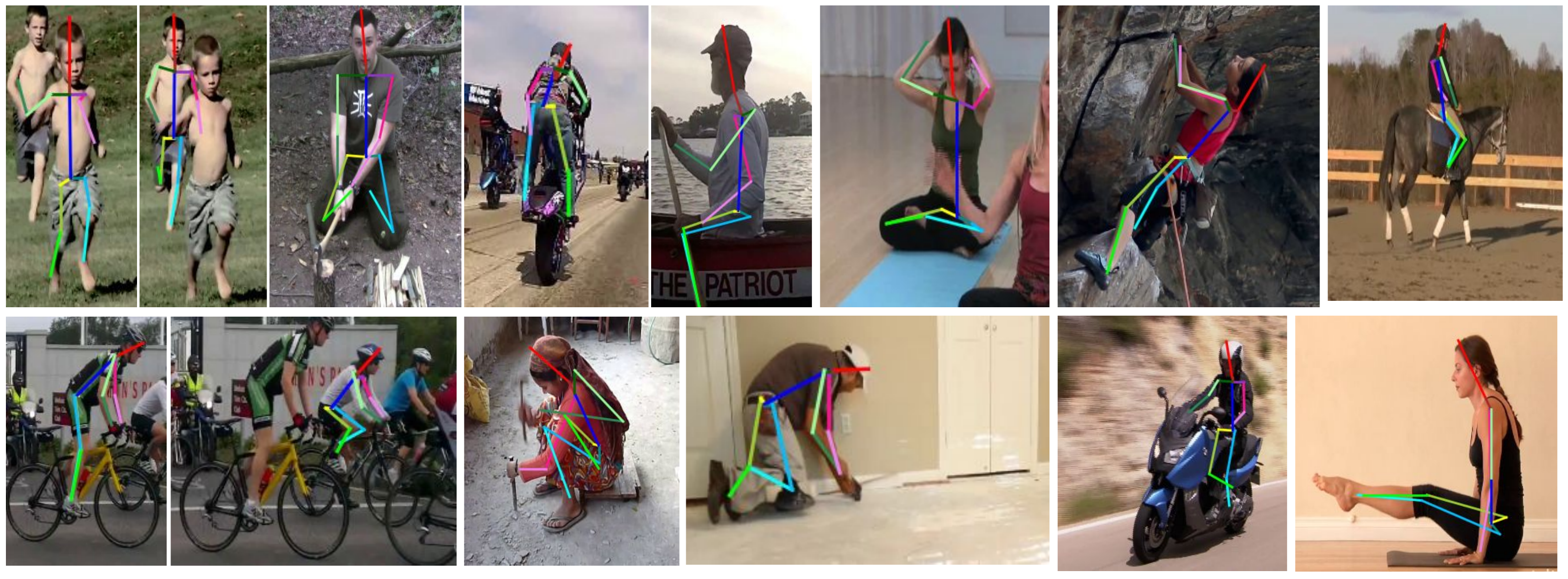}
\caption{Examples of predictions by our chain model on the MPII dataset.}
\label{fig:mpii_examples}
\end{figure}

Fig.~\ref{fig:mpii_examples} shows some examples of our predictions on the MPII dataset.

\subsubsection{Comparison to Other Approaches}
We evaluate our approach on the MPII test set and compare to other methods on the task of pose estimation from a single image. Table~\ref{table:mpii_test} shows the results of our approach and other leading methods in the field. We show the performance of both versions of our chain model, using a shallow 6-layer encoder as well as the inception-v3 architecture. For the shallow chain model, we ensemble two chain models trained at different input scales. For the inception chain model, no ensembling was performed.

The leading approaches by Wei \etal~\cite{wei16} and Newell \etal~\cite{newell16} rely on iteratively refining predictions. In particular, predictions are made initially for all joints  independently. These predictions, which are quite poor (see~\cite{wei16}), are fed subsequently into a network for further refinement. Our approach produces only one set of predictions via a single chain model and does not refine them further. One could combine the two ideas, the one of chained predictions and the one of iterative refinement, to achieve better results.

\setlength{\tabcolsep}{4pt}
\begin{table}[t!]
\begin{center}
\caption{Performance on the MPII test set. A comparison of our chain model, with a shallow 6 layer and an inception-v3 encoder, with leading approaches in the field.}
\label{table:mpii_test}
\begin{tabular}{l|c|c|c|c|c|c|c|c}
Method & Head & Shoulder & Elbow & Wrist & Hip & Knee  & Ankle & Total \\
\hline
\hline
Carreira et al. \cite{IEF2015human}& 95.7  & 91.7  & 81.7  & 72.4  & 82.8  & 73.2 & 66.4 & 81.3  \\
\hline
Tompson et al. \cite{Tompson_2015_CVPR} & 96.1  & 91.9  & 83.9  & 77.8  & 80.9  & 72.3 & 64.8 & 82.0  \\
\hline
Hu\&Ramanan \cite{hu16} & 95.0  & 91.6  & 83.0  & 76.6  & 81.9  & 74.5 & 69.5 & 82.4  \\
\hline
Pishchulin et al. \cite{pischulin16} & 94.1  & 90.2  & 83.4  & 77.3  & 82.6  & 75.7 & 68.6 & 82.4 \\
\hline
Lifshitz et al. \cite{lifshitz16} & 97.8  & 93.3  & 85.7  & 80.4  & 85.3  & 76.6 & 70.2 & 85.0  \\
\hline 
Wei et al. \cite{wei16} & 97.8  & 95.0  & 88.7  & 84.0  & 88.4  & 82.8 & 79.4 & 88.5 \\
\hline
Newell et al. \cite{newell16} & 97.6  & 95.4  & 90.0  & 85.2  & 88.7  & 85.0 & 80.6 & 89.4  \\
\hline
Chain model & 93.8 & 91.8 & 84.2 & 79.4 & 84.4 & 77.9 & 70.7 & 84.1 \\
\hline
\begin{tabular}[c]{@{}l@{}}Inception \\ Chain Model\end{tabular} & 97.9 & 93.2 & 86.7 & 82.1 & 85.2 & 81.5 & 74.0 & 86.1 
\end{tabular}\end{center}
\end{table}
\setlength{\tabcolsep}{1.4pt}

\subsection{Pose Estimation From Videos}
Our chain models in time are described in Equation~\ref{eq:recur_vid} and illustrated in Fig.~\ref{fig:models}. Here, the task
is to localize body parts in time across video frames.  The output variables from the joints of the previous frames are
used as inputs to make a prediction for the joints in the current frame.  We apply the chaining in two different ways
- first, only in time, where each joint is predicted independently of the other joints (as in our baseline models), but
chaining is done in time, and second, with chaining both in time and in joints.

\subsubsection{Pose Estimation in Time}

\setlength{\tabcolsep}{4pt}
\begin{table}[t!]
\begin{center}
	\caption{PCK performance on the Penn Action test set. We show the performance of our chain model for two choices of the time horizon $T_H$ and compare against the per-frame model, with and without temporal smoothing, and a baseline convolutional RNN model. The chain model with $T_H=3$ improves the localization accuracy across all joints. The method by Nie \etal \cite{Nie_2015_CVPR} is shown for comparison. }
\label{table:pose_time}
\begin{tabular}{l|c|c|c|c|c|c|c||c}
PCK  (\%)                                                              & Head     & Shldr      & Elbow         & Wrist         & Hip           & Knee          & Ankle         & Mean       \\
\hline
\hline
\begin{tabular}[c]{@{}l@{}} Nie \etal \cite{Nie_2015_CVPR} \end{tabular} &   64.2    &   55.4       &    33.8      &     24.4     &      56.4      &    54.1    &   48.0      &  48.0 \\
\hline
\begin{tabular}[c]{@{}l@{}}Base Network \end{tabular}    &     94.1    &    90.3     &    84.2      &      83.5     &     88.7      &    87.2    &    87.7     &  87.5 \\
\hline
\begin{tabular}[c]{@{}l@{}}Base Network \\ w/ smoothing \end{tabular} & 93.1  & 91.8 & 85.7   &  78.8   &  90.2   &  91.9 & 91.1 & 88.6 \\
\hline
\begin{tabular}[c]{@{}l@{}}RNN \end{tabular}                  &    95.3   &     92.5    &      87.9      &     87.5      &     91.1       &   89.8     &   90.1      &   90.1 \\
\hline
\begin{tabular}[c]{@{}l@{}}Chain Model, $T_H=1$ \end{tabular}    &  95.8    &      93.2    &       88.9   &      89.6    &     91.3     &      89.8   &     91.2       &  91.0 \\
\hline
\begin{tabular}[c]{@{}l@{}}Chain Model, $T_H=3$ \end{tabular}    &  95.8    &     94.1   &    90.0      &    90.2     &     91.3     &      90.6    &       91.8     &  91.7 \\
\hline
\begin{tabular}[c]{@{}l@{}}Chain Model \\ in time \& joints, $T_H=3$ \end{tabular}    &  95.6    &     93.8   &    90.4      &    90.7     &     91.8     &      90.8    &       91.5     &  91.8 \\
\end{tabular}\end{center}
\end{table}
\setlength{\tabcolsep}{1.4pt}

As shown in Fig.~\ref{fig:models}, the chain model sequentially processes the video frames. The predictions at the previous time steps are used through a recurrent module in order to make a prediction at the current time step. Again, we use a heatmap to encode the location of a part in the frame. 

The details of our learning procedure are identical to the ones described for the single image case. The only difference is that each training example is now a sequence of images $X = \{X_t\}_{t=0}^{T-1}$ each of which has a ground-truth pose. Thus, the loss for $X$ is the sum over the losses for each frame. Each frame loss is defined as in the case of single image (see Eq.~(\ref{eq:single_image_loss})).

We train our model for 120K iterations using SGD with momentum of 0.9, a batch size of 6 and a learning rate of 0.003 with step decay 100K. Images are rescaled to $256\times 256$. A relative radius of $r=0.03$ is used for the loss. The weights are initialized randomly from a normal distribution with zero mean and standard deviation of 0.01.  
 
Table~\ref{table:pose_time} shows the performance on the Penn Action test set. For consistency with previous work on the dataset~\cite{Nie_2015_CVPR}, a prediction is considered correct if it lies within $0.2\times \max(s_h, s_w)$, where $s_h,s_w$ is the height and width, respectively, of the instance in question. We refer to this metric as PCK. (Note that this is a weaker criterion than the one used on the MPII dataset). We show the per frame performance, as produced by a base network $\textrm{CNN}_x$ + $\textrm{CNN}_y$ trained to predict the location of the joints at each frame. We also provide results after applying temporal smoothing to the predictions via the Viterbi algorithm where the transition function is the Euclidean distance of the same joints in two neighboring frames. Additionally, we show the performance of a convolutional RNN with $w_{i,t}^{y}=0$, $\forall i,t$ in Eq.~\ref{eq:recur_vid}. This model corresponds to a standard convolutional RNN where the output variables of the previous time steps are not connected to the hidden state. All networks have roughly the same numbers of parameters, to ensure a fair comparison. For our chain model in time, we show results for two choices of time horizon $T_H$. Namely, $T_H=1$, where predictions of only the previous time step are being considered and $T_H=3$, where predictions of the past 3 frames are considered at each time step. Finally, we show the performance of a chain model in time and in joints, with a time horizon of $T_H=3$.

\begin{figure}[t!]
\centering
\includegraphics[height=5.5cm]{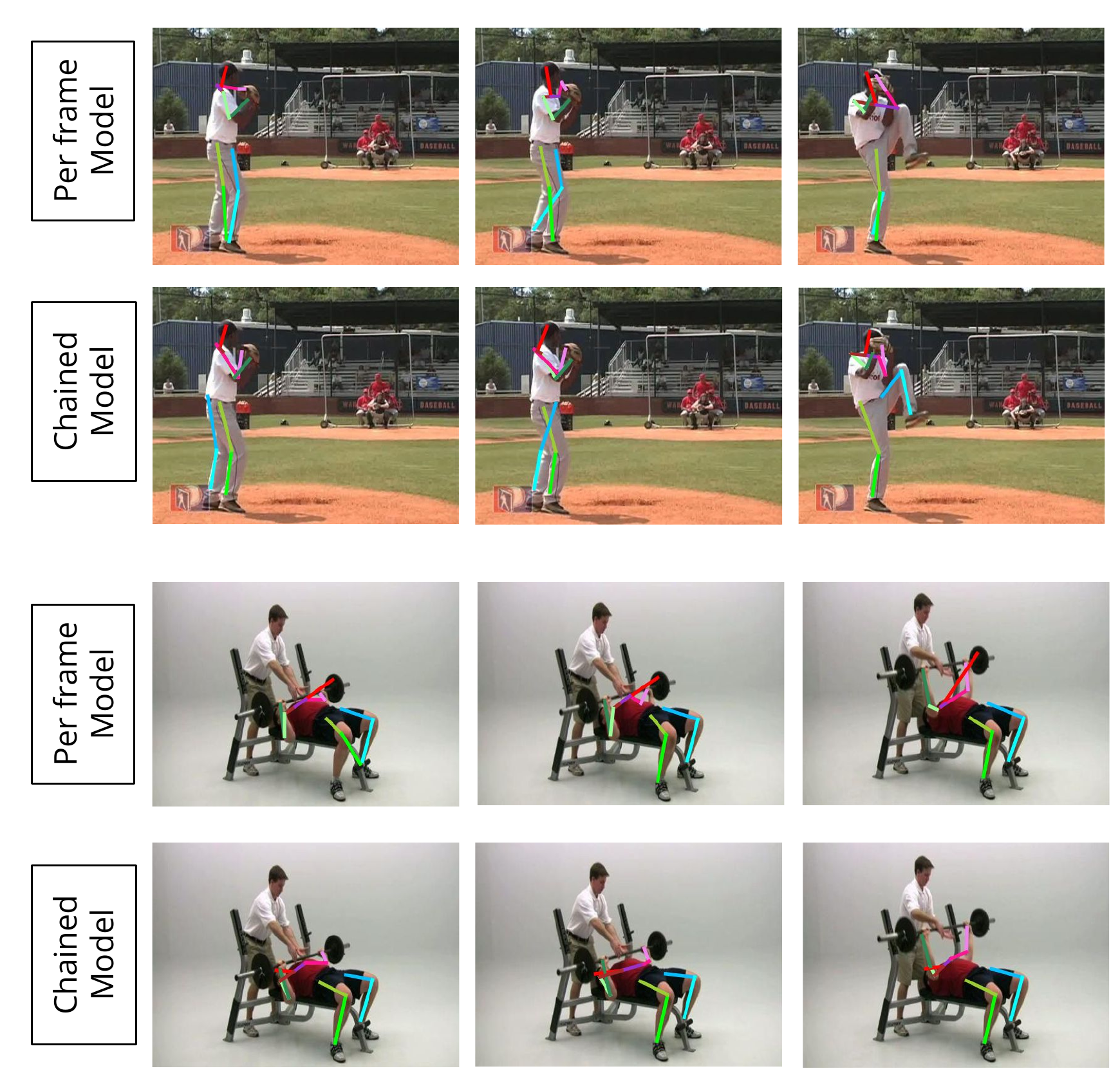}
\includegraphics[height=5.5cm]{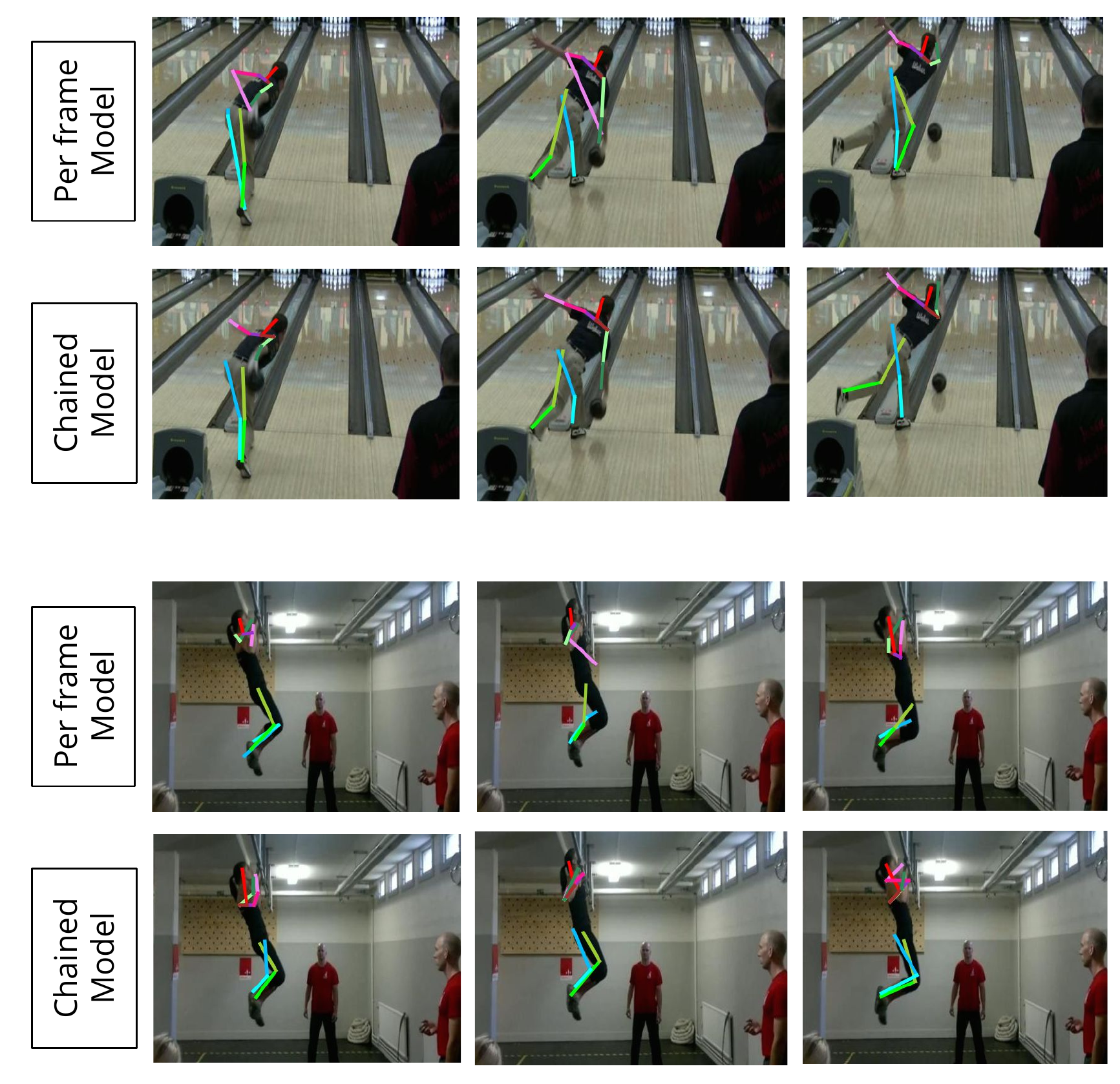}
\caption{Examples of predictions on the Penn Action dataset. Predictions by the per frame model (top) and by the chain model (bottom) are shown in each example block.}
\label{fig:penn_examples}
\end{figure}

We compare to previous work on the Penn Action dataset~\cite{Nie_2015_CVPR}. This model uses action specific pose models, with shallow hand-crafted features, and improves upon Yang \& Ramanan \cite{yang2012articulated}. 

We observe a gain in performance compared to the per frame CNN as well as the RNN across all joints.
Interestingly, chain models show bigger improvement for arms compared to legs. This is due to the fact that
the people in the videos play sports which involve big arm movements, while the legs are mostly un-occluded and less
kinematic. In addition, we see that $T_H=3$ leads to better performance, which is not surprising since the model makes a decision about the location of the joints at the current time step based on observation from 3 past frames. We did not observe additional gains for $T_H>3$. Chaining in time and in joints does not improve performance even further, possibly due to the already high accuracy achieved by the chain model in time.

Fig.~\ref{fig:penn_examples} shows examples of predictions by our chain model on the Penn Action dataset. We also show the predictions made by the per frame detector. We see that the chain model is able to disambiguate right-left confusions which occur often due to the constant
motion of the person while performing actions, while the per frame detector switches very often between erroneous detections.

% Conclusion
%% Conclusion %%
\section{Conclusions}
In this paper, motivated by sequence-to-sequence models, we show how
chained predictions can lead to a powerful tool for structured vision tasks. Chain models allow us to sidestep any assumptions about the joint distribution of the output variables, other than the
capacity of a neural network to model conditional distributions. We prove this point experimentally by showing top performing results on the task of pose estimation from images and videos.
\clearpage

\bibliographystyle{splncs}
\bibliography{refs}
\end{document}